\documentclass[pdflatex,sn-mathphys-num]{sn-jnl}

\usepackage{graphicx}
\usepackage{float}
\usepackage{multirow}
\usepackage{amsmath,amssymb,amsfonts}
\usepackage{amsthm}
\usepackage{mathrsfs}
\usepackage[title]{appendix}
\usepackage{xcolor}
\usepackage{textcomp}
\usepackage{manyfoot}
\usepackage{booktabs}
\usepackage{algorithm}
\usepackage{algorithmicx}
\usepackage{algpseudocode}
\usepackage{listings}

\begin{document}

\title[RAG-TRACE]{RAG-TRACE: Tracing Measurement Evidence and Diagnostic Completeness in AI-Generated ECG Reports}

\author*[1]{\fnm{Elyas} \sur{Irankhah}}\email{Elyas.Irankhah@yale.edu}
\author[2]{\fnm{Shahed} \sur{Salehzehi}}
\author[3]{\fnm{Parisa} \sur{Hosseini}}

\affil*[1]{\orgdiv{Department of Emergency Medicine}, \orgname{Yale University}}

\affil[2]{\orgdiv{Department of Biomedical Engineering}, \orgname{Hakim Sabzevari University}}

\affil[3]{\orgdiv{Neuroscience Research Center}, \orgname{Mashhad University of Medical Sciences}}

\abstract{
\textbf{Purpose:} Retrieval-augmented generation (RAG) can introduce electrocardiogram (ECG) criteria that lack patient-specific measurement support. This study evaluates the reliability of retrieved diagnostic criteria in AI-generated ECG reports.
\textbf{Methods:} We analyzed 1,083 human-validated PTB-XL patient records using a library of 47 criteria cards, including 33 without numeric verification rules. Language models generated structured reports with and without retrieval. Criterion applicability was evaluated using Marquette 12SL measurements, with independent assessment using Uni-G and comparison against expert-validated diagnostic labels.
\textbf{Results:} Keyword retrieval introduced untestable or measurement-disconfirmed criteria in 46.8\% of GPT-4.1-mini reports. A deterministic verification rule withdrew 647 criterion applications, including 639 without numeric checks, while preserving 268 of 269 applications independently supported by Uni-G. Restricting retrieval to 14 testable criteria reduced the initial unsupported-criterion rate to 13.8\%. Under matched instructions and an identical output schema, retrieval increased unsupported-criterion reports from 17.4\% to 31.1\%, while hypertrophy recall remained unchanged at 62.6\%.
\textbf{Conclusion:} These findings identify a fundamental limitation of retrieval-based clinical report generation: retrieved criteria do not necessarily provide valid patient-specific evidence. Measurement-based verification can reduce unsupported criterion applications, but diagnostic validity requires independent assessment.
}

\keywords{electrocardiogram, language models, retrieval, measurement criteria, calibration, PTB-XL}

\maketitle

\section{Introduction}\label{sec1}

The 12-lead electrocardiogram (ECG) is an essential diagnostic tool for assessing cardiac activity through measurable intervals, axes, and voltages \cite{kligfield2007,mason2007}. Large datasets such as PTB-XL, comprising 21,799 recordings from 18,869 patients \cite{wagner2020}, have advanced automated ECG interpretation \cite{strodthoff2021}. Deep learning models have achieved strong performance in detecting cardiac abnormalities \cite{ribeiro2020,hannun2019,alday2020,mehari2022} and supporting cardiovascular screening \cite{attia2019,siontis2021}. Nevertheless, generating reliable clinical reports requires more than diagnostic predictions. Each reported finding must be supported by patient-specific evidence.

Large language models (LLMs) have demonstrated considerable potential for generating clinical narratives \cite{tu2025} and diagnostic explanations \cite{singhal2023,nori2023}. Their outputs can contain unsupported statements and inaccurate evidence attribution \cite{ji2023,maynez2020}. Retrieval-augmented generation (RAG) incorporates external knowledge to improve factual grounding \cite{lewis2020,gao2024survey,shuster2021}, with promising applications in medical question answering \cite{xiong2024,zakka2024} and clinical information extraction \cite{agrawal2022}. Recent studies have investigated evidence alignment in clinical question answering \cite{irankhah2026archehr} and query-level reliability in retriever upgrades \cite{irankhah2026ragflip}. These advances highlight the importance of evidence quality, but the reliability of retrieved criteria in ECG report generation remains insufficiently examined.

ECG interpretation depends on standardized criteria, including QRS duration, Sokolow--Lyon voltage, and corrected QT intervals \cite{surawicz2009,sokolow1949,hancock2009,casale1985,peguero2017,bazett1920,rautaharju2009,prineas2010}. Clinical guidelines establish measurable thresholds and diagnostic definitions \cite{mason2007,wagner2009part6,thygesen2018}. A retrieved criterion may be clinically relevant without being satisfied by the patient's measurements. Other retrieved information, such as calibration settings or laboratory values, may lack a corresponding ECG measurement. These limitations create a critical gap between retrieval relevance and verifiable diagnostic evidence.

This study presents RAG-TRACE, a measurement-based framework for evaluating retrieved criteria in AI-generated ECG reports. We analyze 1,083 held-out PTB-XL patients using 47 criteria cards, including 14 with numeric checks and 33 without measurable verification rules. Structured reports are generated with and without retrieval. A deterministic rule evaluates applied criteria against Marquette 12SL measurements, while Uni-G provides independent assessment. Diagnostic agreement and completeness are evaluated separately against expert-validated labels. Additional experiments assess retrieval constraints, controlled generation settings, and generalizability to other physiological recordings. The evaluation distinguishes criterion applicability, independent measurement agreement, and diagnostic accuracy, following established reporting principles \cite{collins2024,gallifant2025}. Our objective is to determine whether retrieved clinical criteria provide verifiable patient-specific evidence and whether deterministic verification reduces unsupported applications without altering diagnostic conclusions.

\section{Materials and methods}\label{sec2}

\subsection{Data, Measurements, and Diagnostic Probabilities}\label{subsec21}

PTB-XL version 1.0.3 comprises 21,799 ECG recordings from 18,869 patients \cite{wagner2020}. The diagnostic cohort includes 21,375 recordings from 18,610 patients, annotated across five superclasses: normal, myocardial infarction, ST/T change, conduction disturbance, and hypertrophy. Patient-disjoint folds 1--8, 9, and 10 were assigned to training, calibration, and testing, respectively. Sex was encoded as 0 for male and 1 for female. All calibration and test records contained human-validated diagnostic labels, which served as the reference standard without additional clinical adjudication.

ECG measurements were obtained from PTB-XL+ version 1.0.1 \cite{strodthoff2023plus}. Marquette 12SL provided the primary measurements for criterion verification, with Uni-G serving as an independent reference. The 12SL-based verification used Uni-G exclusively for subsequent assessment, while the agreement-based configuration required concordant support from both algorithms. ECGDeli was included in measurement agreement analyses. All measurements were derived from published feature tables, without additional signal delineation or model training.

Diagnostic probabilities for the five superclasses were estimated using one-versus-rest histogram gradient-boosting classifiers trained with inverse-frequency weighting, a learning rate of 0.08, a maximum of 31 leaves, and 120 iterations with early stopping. Class-specific isotonic calibration was performed on fold 9 \cite{niculescu2005}, followed by evaluation on the held-out fold 10. Diagnostic assertions with calibrated probabilities below 0.5 were classified as low-probability, solely for analytical purposes.

\subsection{Criteria cards and retrieval}\label{subsec23}

The criteria library comprised 47 cards, including eight
established measurement thresholds, six opposing numeric
checks, and 33 cards without numeric verification rules.
The original numeric cards use simplified measurement
predicates taken from established ECG standards
\cite{sokolow1949,bazett1920,surawicz2009,
rautaharju2009,hancock2009,prineas2010,wagner2009part6}.
Meeting one predicate is not a clinical diagnosis.
Table~\ref{tab:criteria_library} summarizes the library.

\begin{table}[ht]
\centering
\caption{Composition of the ECG criteria library. Each numeric threshold is an operational predicate used for verification. It is not sufficient, by itself, for a clinical diagnosis.}
\label{tab:criteria_library}
\small
\begin{tabular}{@{}p{0.30\linewidth}p{0.52\linewidth}r@{}}
\toprule
\textbf{Criterion} & \textbf{Threshold or description}
& \textbf{Cards} \\
\midrule
\multicolumn{3}{@{}l}{\textit{Original numeric criteria}} \\
QRS duration & $\geq 120$ ms & 1 \\
PR interval & $\geq 200$ ms & 1 \\
QRS axis & $\leq -30^{\circ}$ & 1 \\
Sokolow--Lyon voltage & $\geq 3.5$ mV & 1 \\
Bazett QTc & $\geq 450$ ms (men);
$\geq 460$ ms (women) & 1 \\
ST deviation (lead II) & $\leq -0.1$ mV & 1 \\
Q-wave duration & $\geq 40$ ms
(II, III, aVF, or V1--V4) & 1 \\
Heart rate & $\geq 100$ bpm & 1 \\
\midrule
Opposing numeric checks & Six complementary
measurement predicates & 6 \\
Cards without numeric checks & Calibration pulse,
paper speed, notch filter, pediatric QRS limit,
lead reversal, age, troponin, wall thickness,
potassium, and digoxin & 33 \\
\midrule
\textbf{Total} & & \textbf{47} \\
\bottomrule
\end{tabular}
\end{table}

\begin{figure*}[ht]
\centering
\includegraphics[width=\textwidth]{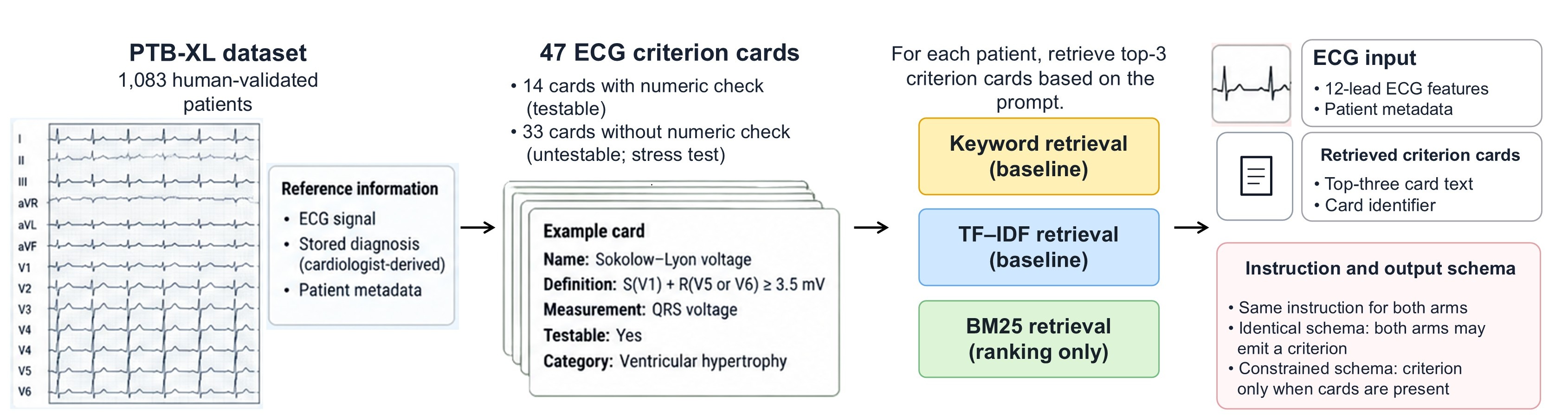}
\caption{Overview of the RAG-TRACE framework, showing dataset preparation, ECG criteria library construction, evidence retrieval strategies, and structured inputs for language model-based ECG report generation.}
\label{fig:workflow}
\end{figure*}

Following construction of the criteria library, three retrieval strategies were evaluated: keyword overlap, term frequency--inverse document frequency (TF--IDF), and Okapi BM25. Retrieval queries were constructed from the names of diagnostic superclasses with calibrated probabilities $\geq 0.5$, without access to held-out diagnostic labels. For keyword retrieval, the relevance score between query $q$ and criterion card $c$ was defined as

\begin{equation}
s(q,c)=\lvert T(q)\cap T(c)\rvert,
\label{eq:overlap}
\end{equation}

where $T(q)$ and $T(c)$ denote their respective token sets. Ties were resolved using card identifiers. TF--IDF and BM25 operated on the same queries and criterion descriptions, with BM25 parameters $k_1=1.5$ and $b=0.75$. Each strategy retrieved the three highest-ranked cards. BM25 was evaluated using saved queries without generating additional reports, and biomedical embedding-based retrieval was not included. Figure~\ref{fig:workflow} illustrates the overall RAG-TRACE pipeline, from dataset preparation and criteria library construction to evidence retrieval and structured ECG report generation.

\subsection{Report Generation and Verification}\label{subsec24}

Structured ECG reports were generated under three conditions: no retrieval, keyword retrieval, and TF--IDF retrieval. GPT-4.1-mini was evaluated at temperature 0, with diagnostic claims restricted to calibrated probabilities $\geq 0.5$. Reports were assessed for diagnostic affirmation using predefined linguistic cues and an exploratory label-aware approach.

\begin{figure*}[ht]
\centering
\includegraphics[width=\textwidth]{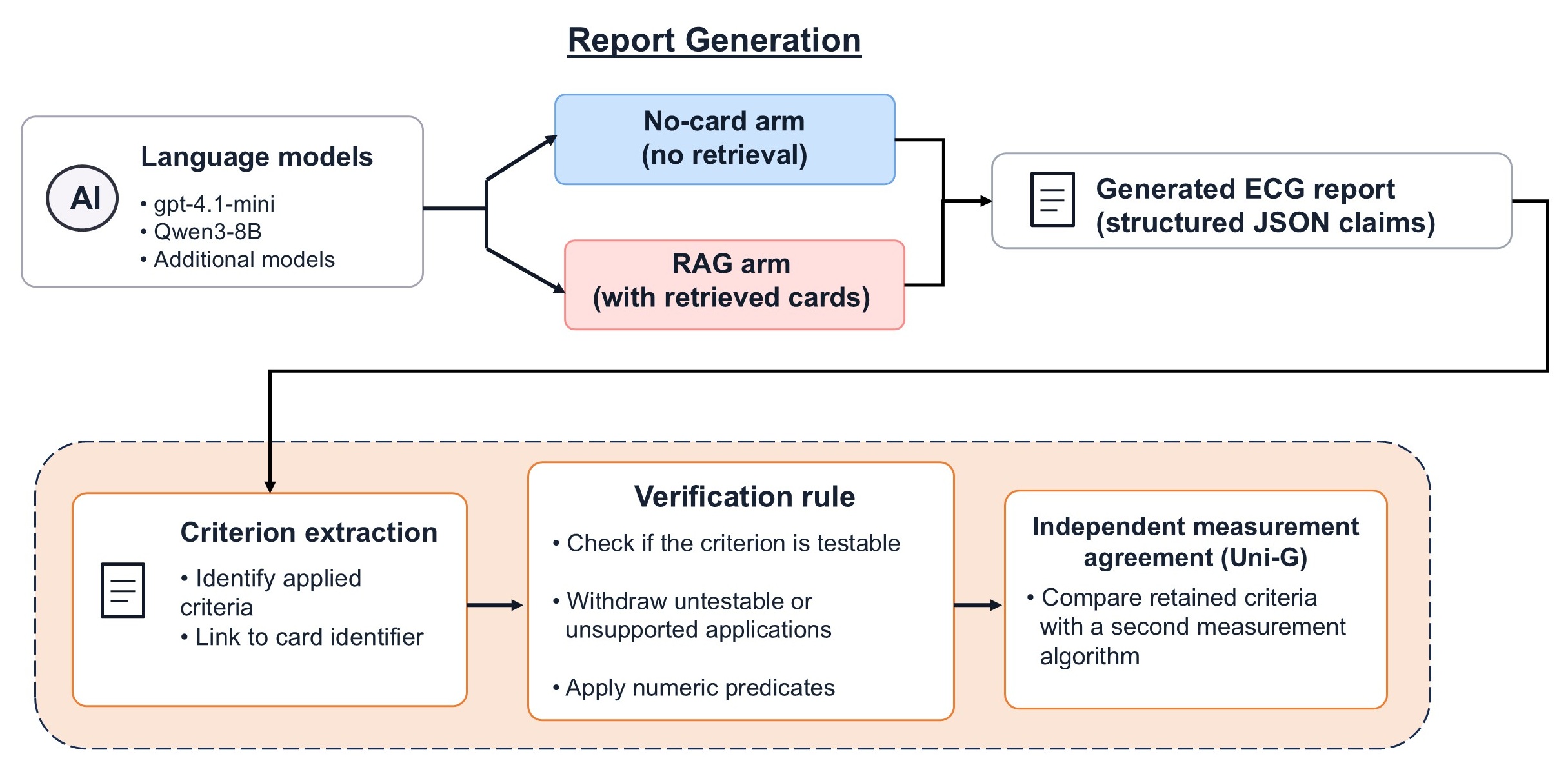}
\caption{RAG-TRACE framework for ECG report generation and measurement-based verification using Marquette 12SL and Uni-G.}
\label{fig:verify}
\end{figure*}

RAG-TRACE applied deterministic verification to retrieved criteria using Marquette 12SL measurements. Criteria without measurable support were withdrawn, while diagnostic statements remained unchanged. Uni-G provided independent assessment, with an additional configuration requiring agreement between both measurement algorithms. The verification process is illustrated in Figure~\ref{fig:verify}.

\begin{figure}[ht]
\centering
\includegraphics[width=\linewidth]{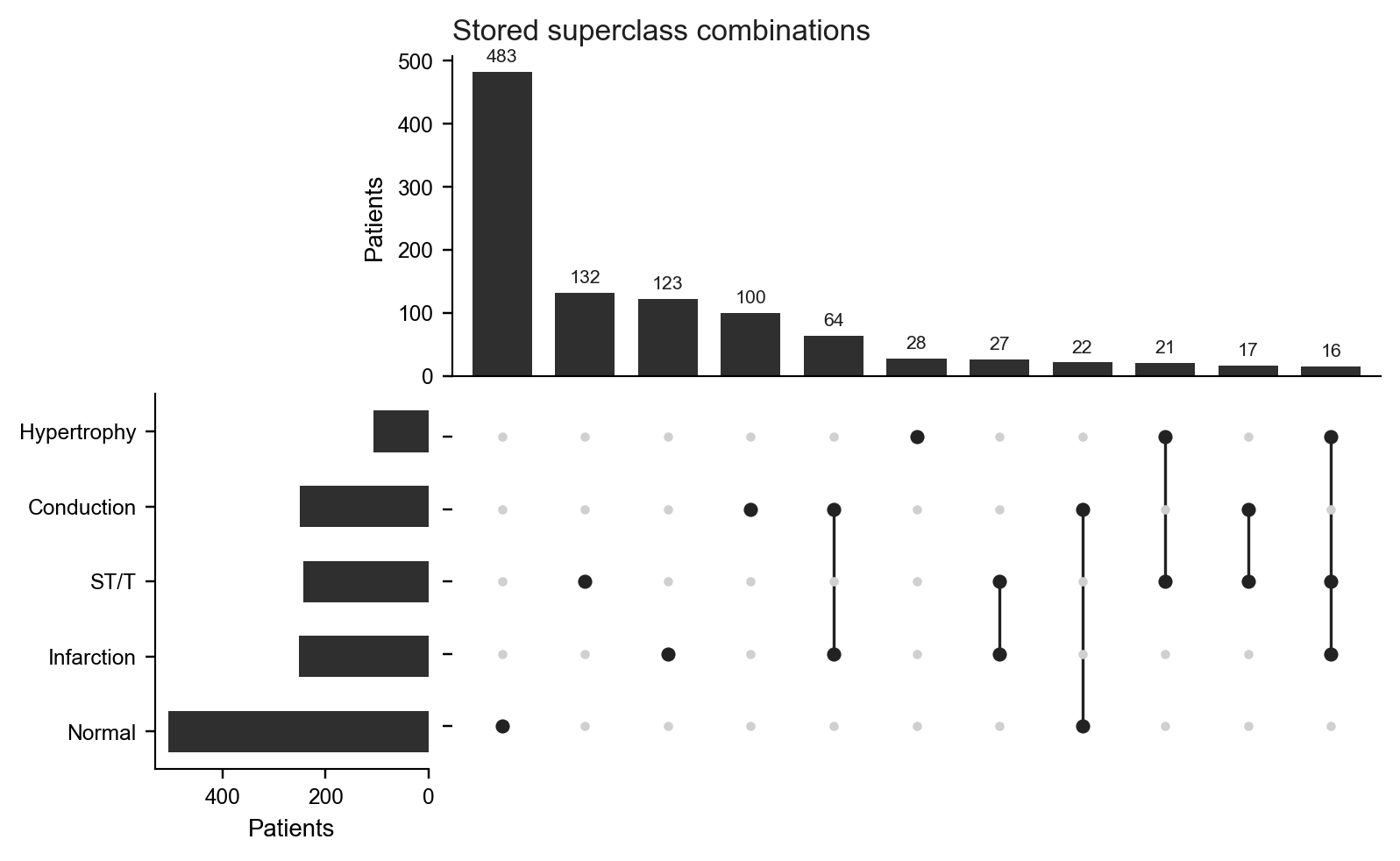}
\caption{Distribution of diagnostic superclass combinations among 1,083 PTB-XL patients. Bars represent combination frequencies, with normal-only recordings forming the largest group ($n=483$).}
\label{fig:upset}
\end{figure}

The evaluation included 1,083 held-out patients, with one ECG per patient. Diagnostic superclass distributions are presented in Figure~\ref{fig:upset}. Statistical comparisons used 10,000 patient-level bootstrap resamples and exact McNemar tests for paired outcomes.

\subsection{Evaluation Metrics}\label{subsec25}

Three complementary metrics were used to assess criterion applicability, independent measurement agreement, and diagnostic consistency with expert-validated labels. For record $i$, let $A_i$ denote the set of applied criteria, $x_i$ the corresponding measurements, and $\kappa(c)$ the numeric verification rule associated with criterion $c$. The predicates $P_{12}$ and $P_U$ represent Marquette 12SL and Uni-G evaluations, respectively. Let $G_i=\{k:y_{i,k}=1\}$ denote the reference diagnostic superclasses and $D_i$ the affirmed superclasses extracted from the generated report. The record-level indicators were defined as

\begin{align}
r_i &= \mathbf{1}\{\exists\, c\in A_i:
\ \kappa(c)=\emptyset \lor \neg P_{12}(c,x_i)\},
\label{eq:release}\\
u_i &= \mathbf{1}\{\exists\, c\in A_i:
\ P_U(c,x_i)=\mathrm{false}\},
\label{eq:unig}\\
e_i &= \mathbf{1}\{D_i\setminus G_i \neq \emptyset\}.
\label{eq:expert}
\end{align}

The corresponding rates were calculated as the mean of each indicator across records. The release score ($r_i$) identifies untestable or measurement-disconfirmed criterion applications. Deterministic 12SL withdrawal sets this score to zero by construction, while the Uni-G score ($u_i$) provides an independent assessment. The agreement-based configuration requires both measurement predicates to hold and consequently sets $u_i=0$ by construction. Withdrawals were categorized by missing numeric checks or failed 12SL predicates, with the latter further classified according to Uni-G agreement. Low-probability diagnostic assertions, defined by calibrated probabilities below 0.5, were evaluated separately. The earlier combined rate represents their union with release-score violations.

Diagnostic consistency ($e_i$) identifies reports affirming at least one superclass absent from the reference labels. Affirmed diagnoses were extracted using the prespecified cue-based rule and, separately, the exploratory label-aware approach. Since verification modifies only criterion applications, diagnostic consistency remains unchanged after withdrawal. An additional statement-level metric assessed discrepancies between the original eight criteria and their linked cardiologist statement codes, including unsupported applications and omitted findings.

Diagnostic agreement was quantified using micro-averaged precision, recall, and F1 across the five superclasses. Class-specific recall was calculated among patients with the corresponding reference diagnosis, while patient-level recall measured the proportion of reference superclasses identified in each report. Paired effects were expressed as $\Delta=\theta_{\mathrm{key}}-\theta_{\mathrm{none}}$, comparing keyword retrieval with no retrieval on the same patients. Confidence intervals were obtained from 10,000 patient-level bootstrap resamples using the 2.5th and 97.5th percentiles.

\subsection{Additional Experiments and Model Evaluation}\label{subsec26}

Additional experiments assessed retrieval constraints, model variability, and robustness. Retrieval was restricted to 14 verifiable criteria, with a separate configuration requiring valid card identifiers. Two GPT-4.1 mini comparisons involving 1,083 patients used fixed instructions with identical or different output schemas to distinguish retrieval effects from schema-dependent diagnostic scores (Figure~\ref{fig:workflow}).

Evaluation included Qwen3-8B and Qwen3-14B \cite{qwen2025}, Gemma 3 12B, MedGemma 4B, DeepSeek-R1-Distill-Llama-8B \cite{deepseek2025}, and Llama 3.1 8B Instruct \cite{llama2024}. Models used greedy decoding and a 1,200-token limit. Paired sample sizes were 1,082, 1,083, 1,083, 850, 112, and 4, respectively; Llama was excluded from rate comparisons. GLM-5.3-Flash was evaluated separately on 1,083 patients per arm.

\section{Results}\label{sec3}

\subsection{Diagnostic Baseline}\label{subsec31}

The calibrated superclass classifier was evaluated on 2,157 held-out ECGs. Macro AUROC, AUPRC, and F1 were 0.935, 0.816, and 0.751, respectively (Table~\ref{tab:baseline}). Isotonic calibration reduced macro expected calibration error from 0.062 to 0.020 \cite{guo2017,naeini2015}. Selective Hamming loss decreased from 0.106 at full coverage to 0.081 at 80\% coverage and 0.047 at 50\% coverage \cite{elyaniv2010,geifman2017}.

\begin{table}[ht]
\caption{Calibrated superclass model on the PTB-XL test fold ($n = 2{,}157$).}\label{tab:baseline}
\begin{tabular}{@{}lccc@{}}
\toprule
Class & AUROC & AUPRC & F1 \\
\midrule
Normal & 0.949 & 0.923 & 0.865 \\
Myocardial infarction & 0.940 & 0.839 & 0.745 \\
ST/T change & 0.934 & 0.793 & 0.750 \\
Conduction disturbance & 0.931 & 0.847 & 0.765 \\
Hypertrophy & 0.918 & 0.680 & 0.633 \\
\midrule
Macro average & 0.935 & 0.816 & 0.751 \\
\bottomrule
\end{tabular}
\end{table}

\subsection{Effects of Criterion Retrieval}\label{subsec32}

All 1,083 GPT-4.1-mini patients and 1,082 Qwen3-8B patients had parseable reports across the three retrieval conditions. For GPT-4.1-mini, the prespecified cue-based expert-absent diagnosis rate increased from 7.5\% without retrieval to 10.1\% with keyword retrieval (paired difference 0.026, 95\% CI 0.006--0.045; McNemar $p=0.008$). TF--IDF produced a rate of 11.0\%. This cue-based metric misses some affirmative statements, as examined in Section~\ref{subsec:labelaware}.

Keyword retrieval produced untestable or 12SL-disconfirmed criterion applications in 46.8\% of GPT-4.1-mini reports. Each report contained an average of 0.891 applied criteria: 0.590 lacked numeric checks, 0.007 failed 12SL predicates, and 0.294 were supported. The release score became zero after deterministic withdrawal by definition; independent Uni-G and expert-label results remained separate (Table~\ref{tab:independent}). At least one top-three card lacked a numeric check in 81.4\% of keyword queries, 79.8\% of BM25 queries, and 73.6\% of TF--IDF queries. 

\begin{table}[ht]
\caption{Keyword retrieval and the same note after the 12SL withdrawal. Expert diagnosis absent: an affirmed superclass is missing from the PTB-XL label. Uni-G rejects: an applied criterion fails the second algorithm. Release: an applied card has no check, or the 12SL predicate is false. The last row is a mean count, not a record rate.}\label{tab:independent}
\begin{tabular}{@{}lcccc@{}}
\toprule
 & \multicolumn{2}{c}{gpt-4.1-mini} & \multicolumn{2}{c}{Qwen3-8B} \\
\cmidrule(lr){2-3}\cmidrule(lr){4-5}
 & Keyword & After 12SL & Keyword & After 12SL \\
\midrule
Expert diagnosis absent & 0.101 & 0.101 & 0.051 & 0.051 \\
Uni-G rejects a criterion & 0.053 & 0.046 & 0.208 & 0.047 \\
Release score & 0.468 & 0 & 0.654 & 0 \\
Mean 12SL-supported criteria kept & 0.294 & 0.294 & 0.306 & 0.306 \\
\bottomrule
\end{tabular}
\end{table}

\subsection{Independent Verification}\label{subsec33}

Following 12SL withdrawal, Uni-G rejection decreased from 5.3\% to 4.6\% for GPT-4.1-mini and from 20.8\% to 4.7\% for Qwen3-8B. Of 647 GPT-4.1 mini withdrawals, 639 involved unchecked cards and eight involved failed numeric predicates. Uni-G independently rejected seven of those eight predicates. For Qwen3-8B, the corresponding counts were 880 unchecked applications and 211 failed predicates, of which Uni-G rejected 203. Verification retained 268 of 269 Uni-G-supported applications for GPT-4.1 mini and 274 of 282 for Qwen3-8B (Table~\ref{tab:retention}). Uni-G was not an input to the 12SL withdrawal rule.

\begin{table}[ht]
\caption{Keyword-note applications judged by Uni-G. Uni-G was not an input to the 12SL edit. The second row is the subset of the failed 12SL predicates that Uni-G also rejected. The unchecked row is separate.}\label{tab:retention}
\begin{tabular}{@{}lcc@{}}
\toprule
 & gpt-4.1-mini & Qwen3-8B \\
\midrule
Uni-G supports, and the edit keeps & 268/269 & 274/282 \\
12SL false, and Uni-G rejects & 7/8 & 203/211 \\
No numeric check, withdrawn & 639/639 & 880/880 \\
12SL predicate false, withdrawn & 8/8 & 211/211 \\
\bottomrule
\end{tabular}
\end{table}

\subsection{Agreement With Expert Statement Codes}\label{subsec34}

For the eight criteria linked to cardiologist statement codes, keyword retrieval increased statement mismatch from 7.5\% to 24.9\% for GPT-4.1 mini and from 1.6\% to 34.1\% for Qwen3-8B. The two-algorithm agreement rule reduced these rates to 22.2\% and 19.6\%, respectively, without a substantial increase in unstated linked findings (Table~\ref{tab:statement}). Among retained applications, the linked code was present in 27.5\% of 204 GPT-4.1 mini applications and 12.8\% of 218 Qwen3-8B applications. Numeric support and diagnostic-label agreement therefore remained distinct.

\begin{table}[ht]
\caption{Statement mismatch against stored cardiologist codes for the original eight checks, and the mean number of those findings left unstated.}\label{tab:statement}
\begin{tabular}{@{}lcccc@{}}
\toprule
 & \multicolumn{2}{c}{gpt-4.1-mini} & \multicolumn{2}{c}{Qwen3-8B} \\
\cmidrule(lr){2-3}\cmidrule(lr){4-5}
Note & Mismatch & Unstated & Mismatch & Unstated \\
\midrule
No retrieval & 0.075 & 0.344 & 0.016 & 0.344 \\
Keyword retrieval & 0.249 & 0.289 & 0.341 & 0.311 \\
12SL withdrawal & 0.246 & 0.289 & 0.228 & 0.314 \\
Both algorithms & 0.222 & 0.293 & 0.196 & 0.318 \\
\bottomrule
\end{tabular}
\end{table}

\subsection{Controlled Retrieval Experiments}\label{subsec35}

Removing the 33 cards without numeric checks reduced the GPT-4.1 mini release score from 46.8\% to 13.8\% (Table~\ref{tab:controlled}). With the 14-card library, requiring a retrieved-card identifier further reduced the release score to 3.0\%. In the identical-schema experiment, both arms used the same instructions and could generate criterion claims; only the keyword arm received retrieved cards. The release score increased from 17.4\% without cards to 31.1\% with cards (Table~\ref{tab:controlled}). Hypertrophy recall was identical in both arms at 0.626 (67 of 107 patients).

\begin{table}[ht]
\caption{Controlled retrieval on the same 1,083 patients. The upper block uses the 14 cards that have a numeric check, and its control column is the original note. The lower block is the identical-schema comparison on the full library. Both arms use one instruction and one JSON schema, and both may emit a criterion claim. Only the keyword arm receives retrieved cards.}\label{tab:controlled}
\begin{tabular}{@{}lccc@{}}
\toprule
 & Control & Keyword & After 12SL \\
\midrule
\multicolumn{4}{@{}l}{\textit{14 numeric cards}} \\
Expert diagnosis absent & 0.075 & 0.098 & 0.098 \\
Uni-G rejects a criterion & 0 & 0.079 & 0.055 \\
Release score & 0 & 0.138 & 0 \\
Mean 12SL-supported criteria kept & 0 & 0.386 & 0.386 \\
\midrule
\multicolumn{4}{@{}l}{\textit{Identical schema, full library}} \\
Expert diagnosis absent & 0.076 & 0.106 & 0.106 \\
Uni-G rejects a criterion & 0 & 0.059 & 0.043 \\
Release score & 0.174 & 0.311 & 0 \\
Mean 12SL-supported criteria kept & 0 & 0.299 & 0.299 \\
\bottomrule
\end{tabular}
\end{table}

\subsection{Model Completion}\label{subsec36}

Llama 3.1 8B Instruct completed all three reports for only four patients and was excluded from rate comparisons. DeepSeek-R1-Distill-Llama-8B contributed 112 complete paired records in the final analysis. An earlier parse contained 116 paired records; its previously reported combined rates were not recomputed on the final subset. These selected samples do not establish full-cohort replication.

\subsection{Waveform and Specificity Controls}\label{subsec37}

The four-head waveform network achieved macro AUROC of 0.981 on the PTB-XL test fold. On 200 MIMIC-IV-ECG studies, AUROC was 0.962 for wide QRS, 0.957 for long PR, 0.943 for left-axis deviation, and 0.998 for tachycardia. In EEG and sleep-recording specificity controls, the withdrawal rule addressed criterion-related errors without correcting unrelated errors.

\subsection{Exploratory Diagnostic Agreement}\label{subsec:labelaware}

\begin{figure}[H]
\centering
\includegraphics[width=\linewidth]{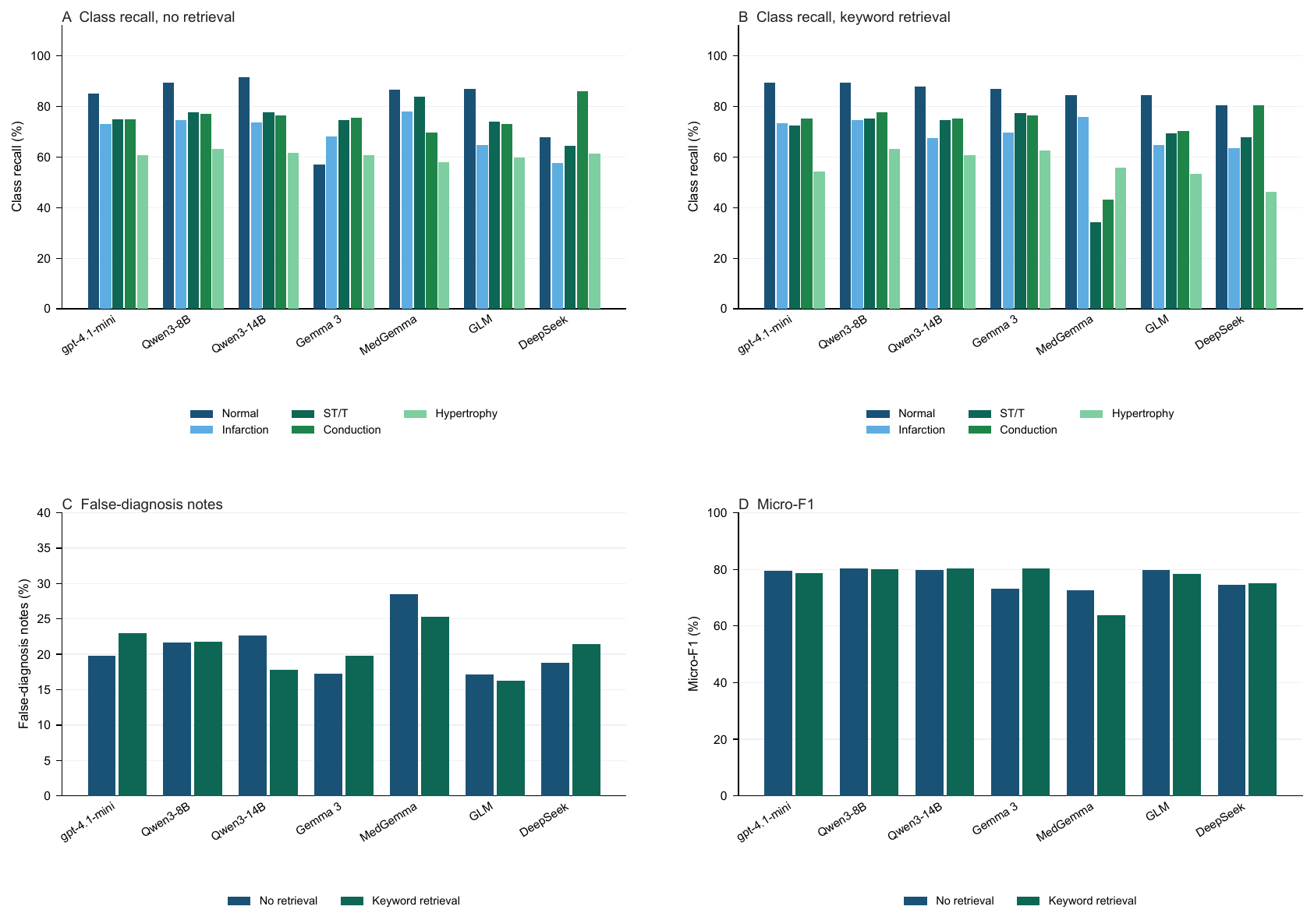}
\caption{Diagnosis agreement for the same models. A, class recall with no retrieval. B, class recall with keyword retrieval. C, notes with a false diagnosis. D, micro-F1. In C and D, blue is no retrieval and green is keyword retrieval. Sample sizes are 1,083 except Qwen3-8B (1,082), MedGemma (850), and DeepSeek (112). DeepSeek is a selected subset.}\label{fig:diagnosis}
\end{figure}

\begin{figure}[H]
\centering
\includegraphics[width=0.72\linewidth]{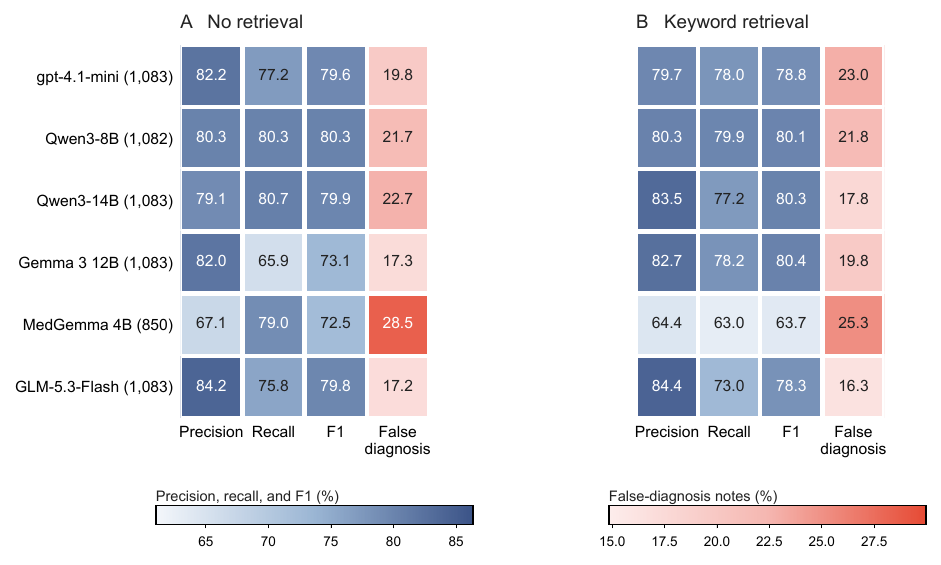}
\caption{Precision, recall, micro-F1, and false-diagnosis notes for the six models with a full paired sample. A, no retrieval. B, keyword retrieval. Cell values are percentages. Blue darkens as precision, recall, and F1 rise. Red darkens as the false-diagnosis rate rises. The row label gives the number of patients. MedGemma is the 850 patients with all three notes. DeepSeek is omitted.}\label{fig:heatmap}
\end{figure}

\begin{figure}[H]
\centering
\includegraphics[width=0.78\linewidth]{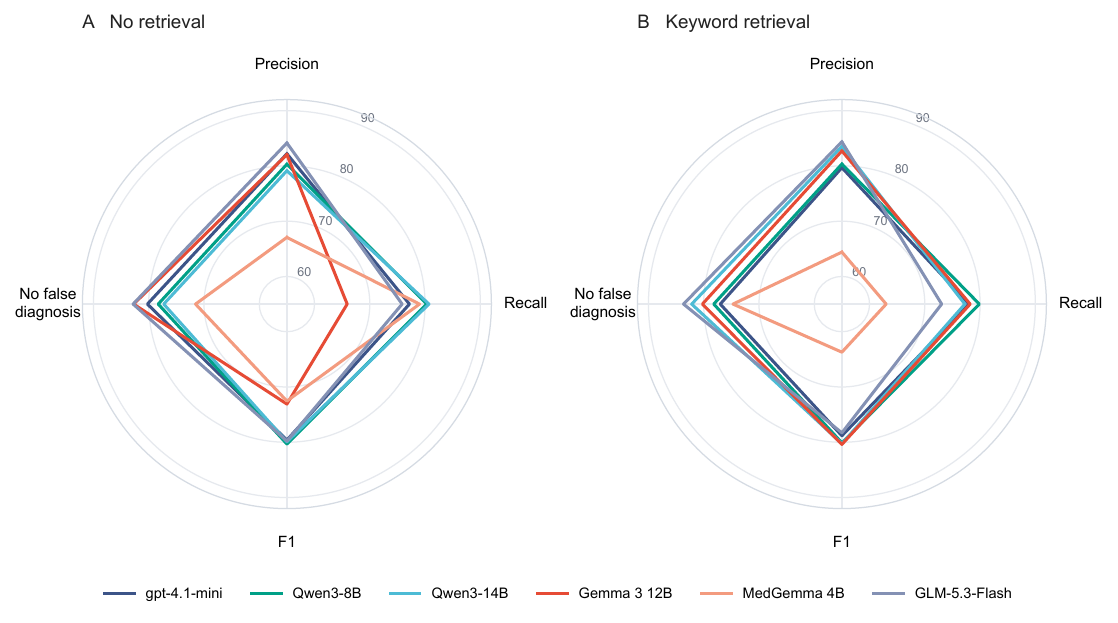}
\caption{The same four measures for those six models. Each axis rises when agreement improves. The fourth axis is the percentage of notes with no false diagnosis. The radial scale starts at 55. A, no retrieval. B, keyword retrieval. MedGemma is the 850 patients with all three notes.}\label{fig:radar}
\end{figure}

The prespecified cue list recognized only 3 of 15 audited ordinary diagnosis sentences and omitted 846 of 1,360 keyword-note diagnosis sentences. An exploratory label-aware rule was therefore evaluated separately. In a 1,000-sentence audit, two language-model judges agreed on 862 sentences. Against their shared labels, the label-aware rule achieved micro-F1 of 0.932, compared with 0.435 for the cue list. Under the label-aware rule, GPT-4.1 mini expert-absent diagnosis rates increased from 19.8\% without cards to 23.0\% with keyword cards (95\% CI for the difference: 0.016--0.049). In the matched-prompt comparison, rates were 21.2\% and 22.1\%, with a confidence interval including zero. Effects differed across models: Qwen3-14B and MedGemma 4B showed lower expert-absent rates with retrieval, Gemma 3 12B showed a higher rate, and Qwen3-8B showed little change (Table~\ref{tab:labelaware}). Figures~\ref{fig:diagnosis}, \ref{fig:heatmap}, and \ref{fig:radar} summarize class recall, diagnostic precision, micro-F1, and expert-absent diagnosis rates across the saved original-prompt comparisons.

\begin{table}[ht]
\caption{Exploratory label-aware expert-absent rates. A counted diagnosis is a saved superclass label that the sentence does not negate. The no-card column for the numeric-only and constrained arms is the original no-card note, joined by electrocardiogram identifier. Intervals are 95\% bootstrap confidence intervals.}\label{tab:labelaware}
\begin{tabular}{@{}lcccc@{}}
\toprule
Arm & $n$ & No cards & Keyword & Difference \\
\midrule
gpt-4.1-mini, full library & 1,083 & 0.198 & 0.230 & 0.032 (0.016, 0.049) \\
gpt-4.1-mini, 14 numeric cards & 1,083 & 0.198 & 0.252 & 0.054 (0.038, 0.072) \\
gpt-4.1-mini, matched prompt & 1,083 & 0.212 & 0.221 & 0.008 ($-0.001$, 0.018) \\
gpt-4.1-mini, constrained identifier & 1,083 & 0.198 & 0.140 & $-0.057$ ($-0.077$, $-0.038$) \\
Qwen3-8B & 1,082 & 0.217 & 0.218 & 0.001 ($-0.009$, 0.011) \\
Qwen3-14B & 1,083 & 0.227 & 0.178 & $-0.049$ ($-0.062$, $-0.037$) \\
Gemma 3 12B & 1,083 & 0.173 & 0.198 & 0.025 (0.015, 0.036) \\
MedGemma 4B & 850 & 0.285 & 0.253 & $-0.032$ ($-0.055$, $-0.007$) \\
DeepSeek-R1 8B & 112 & 0.188 & 0.214 & 0.027 ($-0.018$, 0.071) \\
\bottomrule
\end{tabular}
\end{table}

\subsection{Schema Effects on Diagnostic Agreement}\label{subsec:diagnosis}

\begin{figure}[H]
\centering
\includegraphics[width=\linewidth]{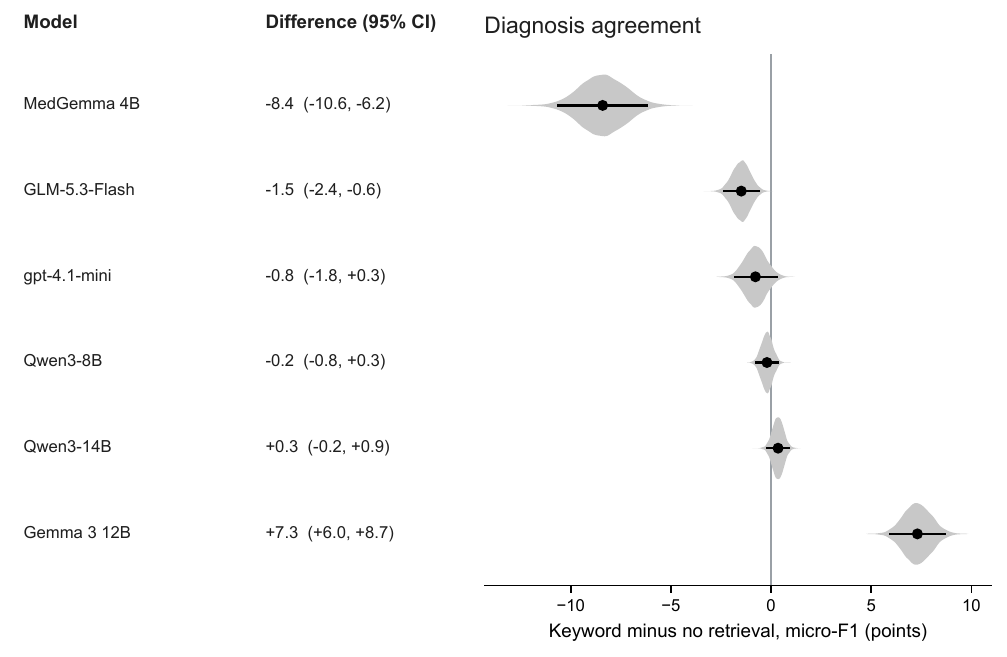}
\caption{Diagnosis micro-F1 change under keyword retrieval. Each row is keyword minus no retrieval on the same patients. The dot is the estimate, the line is the 95\% bootstrap interval, and the gray shape is the bootstrap distribution. The interval uses 10,000 patient resamples. MedGemma is the 904 patients who have both notes. The three-note table uses 850.}\label{fig:forest}
\end{figure}

A separate different-schema comparison used the same instructions but allowed criterion claims only when retrieved cards were present. Micro-F1 decreased from 0.804 to 0.795 (difference $-0.009$, 95\% CI $-0.014$ to $-0.004$), while scored hypertrophy recall decreased from 0.617 to 0.514 (Table~\ref{tab:matched}). The expert-absent diagnosis rate changed from 17.5\% to 16.7\% ($p=0.077$). Of 11 scored hypertrophy losses, six no-card credits involved a stored HYP label attached to hypertension wording. Other discrepancies involved denied voltage criteria, qualified diagnoses, or endorsements outside diagnosis claims. These scores do not represent 11 clinically missed diagnoses. The identical-schema comparison showed unchanged hypertrophy recall. An earlier identifier-constrained comparison joined against a different no-card prompt showed a larger recall decline (0.772 to 0.546), which cannot be attributed solely to retrieval. Figure~\ref{fig:forest} presents paired micro-F1 changes across models. All diagnosis comparisons use stored PTB-XL codes, not newly adjudicated clinician interpretations.

\begin{table}[ht]
\caption{Different-schema diagnosis agreement for gpt-4.1-mini, $n = 1{,}083$. Both arms use one instruction. The schema admits a criterion claim only when retrieved cards are present. The identical-schema comparison is Table~\ref{tab:controlled}. Precision, recall, and F1 are micro averages against the stored PTB-XL superclass codes. A false-diagnosis note affirms at least one superclass that is absent from that code.}\label{tab:matched}
\begin{tabular}{@{}lccccc@{}}
\toprule
Arm & Precision & Recall & F1 & False-diagnosis notes & Omissions \\
\midrule
No cards & 0.841 & 0.770 & 0.804 & 0.175 & 312 \\
14-card constraint & 0.844 & 0.752 & 0.795 & 0.167 & 337 \\
\bottomrule
\end{tabular}
\end{table}

An exploratory classification of 965 GPT-4.1 mini keyword applications identified 639 without a numeric check, 174 supported, 150 indeterminate, and two refuted. Among the original eight linked checks, the corresponding cardiologist code was present for 34.8\% of 132 supported applications and 12.0\% of 117 indeterminate applications. 


\section{Discussion}\label{sec4}

Retrieval-augmented ECG reporting exposes a gap between clinical relevance and patient-specific evidence. Keyword retrieval introduced untestable or measurement-disconfirmed criteria in 46.8\% of GPT-4.1 mini reports, with 639 of 647 withdrawn applications lacking numeric checks. Qwen3-8B exhibited a distinct failure pattern, with 203 of 211 measurement-disconfirmed applications independently rejected by Uni-G. The differences highlight two sources of unreliable reporting: unverifiable retrieved content and incorrect application of measurable criteria. The deliberately constructed criteria library increased exposure to these failures, limiting generalization to routine clinical practice. RAG-TRACE complements research on retrieval robustness, medical citation reliability, and multi-step clinical reasoning \cite{parklee2024robust,wu2025sourcecheckup,wind2025radiology} by assessing retrieved evidence against patient-specific measurements. Verification preserved 268 of 269 independently supported GPT-4.1 mini applications and 274 of 282 Qwen3-8B applications. This selective withdrawal offers that unsupported criteria can be removed with limited loss of measurement-supported claims. The zero release score reflects the deterministic rule itself, making independent measurement agreement essential for assessing verification performance.

Controlled comparisons further distinguish retrieval effects from changes in diagnostic scoring. Under identical instructions and output schemas, unsupported-criterion reports increased from 17.4\% to 31.1\%, while hypertrophy recall remained at 62.6\%. The different-schema comparison showed a modest decrease in micro-F1 (0.804 to 0.795) and scored hypertrophy recall (0.617 to 0.514). Six of the 11 apparent hypertrophy losses involved hypertension statements assigned hypertrophy labels. Such discrepancies illustrate how output structure and scoring definitions can influence apparent diagnostic performance.

For clinical RAG systems, evidence retrieval and criterion verification serve distinct purposes. A clinically relevant criterion requires measurable patient support before its application can be justified. Deterministic verification can address unsupported criterion claims without modifying diagnostic statements. Measurement agreement, diagnostic completeness, and clinical validity must therefore be assessed independently.

\section{Limitations}\label{subsec44}

The evaluation relied on published PTB-XL diagnostic labels without independent clinical review of generated reports. The prespecified affirmation rule undercounted diagnostic assertions, while the exploratory label-aware analysis excluded 138 sentences with disagreement between judges. Model comparisons were constrained by incomplete JSON generation for Llama, 112 paired DeepSeek cases, and the use of a single closed model for numeric-only retrieval. The identical-schema and different-schema comparisons addressed distinct experimental conditions. BM25 was evaluated without report generation, and TF--IDF did not incorporate biomedical embeddings. The reported cue-list contrast was significant ($p = 0.008$); the other comparisons were exploratory. The absence of prospective clinical evaluation limits conclusions about diagnostic safety and real-world applicability.

\section{Conclusion}\label{sec5}

RAG-TRACE establishes a measurement-centered approach to assessing evidence reliability in AI-generated ECG reports. The evaluation highlights the importance of distinguishing retrieved clinical information from criteria supported by patient-specific measurements. Its verification mechanism provides a transparent means of identifying unsupported applications while preserving valid quantitative evidence. Importantly, criterion-level reliability and diagnostic completeness represent separate dimensions of report quality that require independent assessment. Integrating these principles into clinical RAG pipelines could support more interpretable and auditable AI-generated reports. Future work should extend RAG-TRACE to broader clinical datasets, incorporate measurement-aware retrieval, and evaluate its impact through clinician-adjudicated assessments.

\section*{Declarations}

\textbf{Funding.} Not applicable.

\textbf{Competing interests.} The authors declare no competing interests.

\textbf{Ethics approval.} Not applicable. The study used public recordings and recruited no participants.

\textbf{Author contributions.} All authors contributed to the manuscript and approved the submitted version.

\textbf{Data and code availability.} All code and study materials are publicly available at \url{https://github.com/Elyasirankhah/RAG-TRACE}.

\bibliography{sn-bibliography}

@article{wagner2020,
  author  = {Wagner, Patrick and Strodthoff, Nils and Bousseljot, Ralf-Dieter and Kreiseler, Dieter and Lunze, Fatima I. and Samek, Wojciech and Schaeffter, Tobias},
  title   = {{PTB-XL}, a large publicly available electrocardiography dataset},
  journal = {Scientific Data},
  year    = {2020},
  volume  = {7},
  pages   = {154},
  doi     = {10.1038/s41597-020-0495-6}
}

@article{strodthoff2021,
  author  = {Strodthoff, Nils and Wagner, Patrick and Schaeffter, Tobias and Samek, Wojciech},
  title   = {Deep learning for {ECG} analysis: benchmarks and insights from {PTB-XL}},
  journal = {{IEEE} Journal of Biomedical and Health Informatics},
  year    = {2021},
  volume  = {25},
  number  = {5},
  pages   = {1519--1528},
  doi     = {10.1109/JBHI.2020.3022989}
}

@article{strodthoff2023plus,
  author  = {Strodthoff, Nils and Mehari, Temesgen and Nagel, Claudia and Aston, Philip J. and Sundar, Ashish and Graff, Claus and Kanters, J{\o}rgen K. and Haverkamp, Wilhelm and D{\"o}ssel, Olaf and Loewe, Axel and B{\"a}r, Markus and Schaeffter, Tobias},
  title   = {{PTB-XL+}, a comprehensive electrocardiographic feature dataset},
  journal = {Scientific Data},
  year    = {2023},
  volume  = {10},
  pages   = {279},
  doi     = {10.1038/s41597-023-02153-8}
}

@article{sokolow1949,
  author  = {Sokolow, Maurice and Lyon, Thomas P.},
  title   = {The ventricular complex in left ventricular hypertrophy as obtained by unipolar precordial and limb leads},
  journal = {American Heart Journal},
  year    = {1949},
  volume  = {37},
  number  = {2},
  pages   = {161--186}
}

@article{bazett1920,
  author  = {Bazett, Henry C.},
  title   = {An analysis of the time-relations of electrocardiograms},
  journal = {Heart},
  year    = {1920},
  volume  = {7},
  pages   = {353--370}
}

@article{kligfield2007,
  author  = {Kligfield, Paul and Gettes, Leonard S. and Bailey, James J. and Childers, Rory and Deal, Barbara J. and Hancock, E. William and van Herpen, Gerard and Kors, Jan A. and Macfarlane, Peter and Mirvis, David M. and others},
  title   = {Recommendations for the standardization and interpretation of the electrocardiogram. {Part I}: the electrocardiogram and its technology},
  journal = {Circulation},
  year    = {2007},
  volume  = {115},
  number  = {10},
  pages   = {1306--1324}
}

@article{mason2007,
  author  = {Mason, Jay W. and Hancock, E. William and Gettes, Leonard S.},
  title   = {Recommendations for the standardization and interpretation of the electrocardiogram. {Part II}: electrocardiography diagnostic statement list},
  journal = {Circulation},
  year    = {2007},
  volume  = {115},
  number  = {10},
  pages   = {1325--1332}
}

@article{surawicz2009,
  author  = {Surawicz, Borys and Childers, Rory and Deal, Barbara J. and Gettes, Leonard S.},
  title   = {Recommendations for the standardization and interpretation of the electrocardiogram. {Part III}: intraventricular conduction disturbances},
  journal = {Circulation},
  year    = {2009},
  volume  = {119},
  number  = {10},
  pages   = {e235--e240}
}

@article{rautaharju2009,
  author  = {Rautaharju, Pentti M. and Surawicz, Borys and Gettes, Leonard S.},
  title   = {Recommendations for the standardization and interpretation of the electrocardiogram. {Part IV}: the {ST} segment, {T} and {U} waves, and the {QT} interval},
  journal = {Circulation},
  year    = {2009},
  volume  = {119},
  number  = {10},
  pages   = {e241--e250}
}

@article{hancock2009,
  author  = {Hancock, E. William and Deal, Barbara J. and Mirvis, David M. and Okin, Peter and Kligfield, Paul and Gettes, Leonard S.},
  title   = {Recommendations for the standardization and interpretation of the electrocardiogram. {Part V}: electrocardiogram changes associated with cardiac chamber hypertrophy},
  journal = {Circulation},
  year    = {2009},
  volume  = {119},
  number  = {10},
  pages   = {e251--e261}
}

@article{wagner2009part6,
  author  = {Wagner, Galen S. and Macfarlane, Peter and Wellens, Hein and Josephson, Mark and Gorgels, Anton and Mirvis, David M. and Pahlm, Olle and Surawicz, Borys and Kligfield, Paul and Childers, Rory and others},
  title   = {Recommendations for the standardization and interpretation of the electrocardiogram. {Part VI}: acute ischemia/infarction},
  journal = {Circulation},
  year    = {2009},
  volume  = {119},
  number  = {10},
  pages   = {e262--e270}
}

@article{thygesen2018,
  author  = {Thygesen, Kristian and Alpert, Joseph S. and Jaffe, Allan S. and Chaitman, Bernard R. and Bax, Jeroen J. and Morrow, David A. and White, Harvey D.},
  title   = {Fourth universal definition of myocardial infarction (2018)},
  journal = {Circulation},
  year    = {2018},
  volume  = {138},
  number  = {20},
  pages   = {e618--e651}
}

@article{ribeiro2020,
  author  = {Ribeiro, Ant{\^o}nio H. and Ribeiro, Manoel Horta and Paix{\~a}o, Gabriela M. M. and Oliveira, Derick M. and Gomes, Paulo R. and Canazart, J{\'e}ssica A. and Ferreira, Milton P. S. and Andersson, Carl R. and Macfarlane, Peter W. and Meira Jr, Wagner and others},
  title   = {Automatic diagnosis of the 12-lead {ECG} using a deep neural network},
  journal = {Nature Communications},
  year    = {2020},
  volume  = {11},
  pages   = {1760}
}

@article{hannun2019,
  author  = {Hannun, Awni Y. and Rajpurkar, Pranav and Haghpanahi, Masoumeh and Tison, Geoffrey H. and Bourn, Codie and Turakhia, Mintu P. and Ng, Andrew Y.},
  title   = {Cardiologist-level arrhythmia detection and classification in ambulatory electrocardiograms using a deep neural network},
  journal = {Nature Medicine},
  year    = {2019},
  volume  = {25},
  pages   = {65--69}
}

@article{attia2019,
  author  = {Attia, Zachi I. and Kapa, Suraj and Lopez-Jimenez, Francisco and McKie, Paul M. and Ladewig, Dorothy J. and Satam, Gaurav and Pellikka, Patricia A. and Enriquez-Sarano, Maurice and Noseworthy, Peter A. and Munger, Thomas M. and others},
  title   = {Screening for cardiac contractile dysfunction using an artificial intelligence-enabled electrocardiogram},
  journal = {Nature Medicine},
  year    = {2019},
  volume  = {25},
  pages   = {70--74},
  doi     = {10.1038/s41591-018-0240-2}
}

@article{siontis2021,
  author  = {Siontis, Konstantinos C. and Noseworthy, Peter A. and Attia, Zachi I. and Friedman, Paul A.},
  title   = {Artificial intelligence-enhanced electrocardiography in cardiovascular disease management},
  journal = {Nature Reviews Cardiology},
  year    = {2021},
  volume  = {18},
  pages   = {465--478}
}

@article{alday2020,
  author  = {Alday, Erick A. Perez and Gu, Annie and Shah, Amit J. and Robichaux, Chad and Wong, An-Kwok Ian and Liu, Chengyu and Liu, Feifei and Rad, Ali Bahrami and Elola, Andoni and Seyedi, Salman and others},
  title   = {Classification of 12-lead {ECG}s: the {PhysioNet}/{Computing in Cardiology} Challenge 2020},
  journal = {Physiological Measurement},
  year    = {2020},
  volume  = {41},
  number  = {12},
  pages   = {124003},
  doi     = {10.1088/1361-6579/abc960}
}

@inproceedings{lewis2020,
  author    = {Lewis, Patrick and Perez, Ethan and Piktus, Aleksandra and Petroni, Fabio and Karpukhin, Vladimir and Goyal, Naman and K{\"u}ttler, Heinrich and Lewis, Mike and Yih, Wen-tau and Rockt{\"a}schel, Tim and others},
  title     = {Retrieval-augmented generation for knowledge-intensive {NLP} tasks},
  booktitle = {Advances in Neural Information Processing Systems},
  year      = {2020},
  volume    = {33},
  pages     = {9459--9474}
}

@inproceedings{shuster2021,
  author    = {Shuster, Kurt and Poff, Spencer and Chen, Moya and Kiela, Douwe and Weston, Jason},
  title     = {Retrieval augmentation reduces hallucination in conversation},
  booktitle = {Findings of the Association for Computational Linguistics: {EMNLP}},
  year      = {2021},
  pages     = {3784--3803}
}

@article{gao2024survey,
  author  = {Gao, Yunfan and Xiong, Yun and Gao, Xinyu and Jia, Kangxiang and Pan, Jinliu and Bi, Yuxi and Dai, Yi and Sun, Jiawei and Wang, Meng and Wang, Haofen},
  title   = {Retrieval-augmented generation for large language models: a survey},
  journal = {arXiv preprint arXiv:2312.10997},
  year    = {2024}
}

@inproceedings{xiong2024,
  author    = {Xiong, Guangzhi and Jin, Qiao and Lu, Zhiyong and Zhang, Aidong},
  title     = {Benchmarking retrieval-augmented generation for medicine},
  booktitle = {Findings of the Association for Computational Linguistics: {ACL}},
  year      = {2024},
  pages     = {6233--6251}
}

@article{zakka2024,
  author  = {Zakka, Cyril and Shad, Rohan and Chaurasia, Akash and Dalal, Alex R. and Kim, Jennifer L. and Moor, Michael and Fong, Robyn and Phillips, Curran and Alexander, Kevin and Ashley, Euan and others},
  title   = {Almanac---retrieval-augmented language models for clinical medicine},
  journal = {{NEJM AI}},
  year    = {2024},
  volume  = {1},
  number  = {2}
}

@article{singhal2023,
  author  = {Singhal, Karan and Azizi, Shekoofeh and Tu, Tao and Mahdavi, S. Sara and Wei, Jason and Chung, Hyung Won and Scales, Nathan and Tanwani, Ajay and Cole-Lewis, Heather and Pfohl, Stephen and others},
  title   = {Large language models encode clinical knowledge},
  journal = {Nature},
  year    = {2023},
  volume  = {620},
  pages   = {172--180}
}

@article{nori2023,
  author  = {Nori, Harsha and King, Nicholas and McKinney, Scott Mayer and Carignan, Dean and Horvitz, Eric},
  title   = {Capabilities of {GPT-4} on medical challenge problems},
  journal = {arXiv preprint arXiv:2303.13375},
  year    = {2023}
}

@article{tu2025,
  author  = {Tu, Tao and Schaekermann, Mike and Palepu, Anil and Saab, Khaled and Freyberg, Jan and Tanno, Ryutaro and Wang, Amy and Li, Brenna and Amin, Mohamed and Cheng, Yong and others},
  title   = {Towards conversational diagnostic artificial intelligence},
  journal = {Nature},
  year    = {2025},
  volume  = {642},
  pages   = {442--450}
}

@article{ji2023,
  author  = {Ji, Ziwei and Lee, Nayeon and Frieske, Rita and Yu, Tiezheng and Su, Dan and Xu, Yan and Ishii, Etsuko and Bang, Ye Jin and Madotto, Andrea and Fung, Pascale},
  title   = {Survey of hallucination in natural language generation},
  journal = {{ACM} Computing Surveys},
  year    = {2023},
  volume  = {55},
  number  = {12},
  pages   = {1--38}
}

@inproceedings{maynez2020,
  author    = {Maynez, Joshua and Narayan, Shashi and Bohnet, Bernd and McDonald, Ryan},
  title     = {On faithfulness and factuality in abstractive summarization},
  booktitle = {Proceedings of the 58th Annual Meeting of the Association for Computational Linguistics},
  year      = {2020},
  pages     = {1906--1919}
}

@inproceedings{guo2017,
  author    = {Guo, Chuan and Pleiss, Geoff and Sun, Yu and Weinberger, Kilian Q.},
  title     = {On calibration of modern neural networks},
  booktitle = {Proceedings of the 34th International Conference on Machine Learning},
  year      = {2017},
  pages     = {1321--1330}
}

@inproceedings{naeini2015,
  author    = {Naeini, Mahdi Pakdaman and Cooper, Gregory F. and Hauskrecht, Milos},
  title     = {Obtaining well calibrated probabilities using {Bayesian} binning},
  booktitle = {Proceedings of the {AAAI} Conference on Artificial Intelligence},
  year      = {2015},
  volume    = {29},
  number    = {1}
}

@inproceedings{niculescu2005,
  author    = {Niculescu-Mizil, Alexandru and Caruana, Rich},
  title     = {Predicting good probabilities with supervised learning},
  booktitle = {Proceedings of the 22nd International Conference on Machine Learning},
  year      = {2005},
  pages     = {625--632}
}

@article{elyaniv2010,
  author  = {El-Yaniv, Ran and Wiener, Yair},
  title   = {On the foundations of noise-free selective classification},
  journal = {Journal of Machine Learning Research},
  year    = {2010},
  volume  = {11},
  pages   = {1605--1641}
}

@inproceedings{geifman2017,
  author    = {Geifman, Yonatan and El-Yaniv, Ran},
  title     = {Selective classification for deep neural networks},
  booktitle = {Advances in Neural Information Processing Systems},
  year      = {2017},
  volume    = {30}
}

@article{collins2024,
  author  = {Collins, Gary S. and Moons, Karel G. M. and Dhiman, Paula and Riley, Richard D. and Beam, Andrew L. and Van Calster, Ben and Ghassemi, Marzyeh and Liu, Xiaoxuan and Reitsma, Johannes B. and van Smeden, Maarten and others},
  title   = {{TRIPOD+AI} statement: updated guidance for reporting clinical prediction models that use regression or machine learning methods},
  journal = {{BMJ}},
  year    = {2024},
  volume  = {385},
  pages   = {e078378}
}

@article{gallifant2025,
  author  = {Gallifant, Jack and Afshar, Majid and Ameen, Saleem and Aphinyanaphongs, Yindalon and Chen, Shan and Cacciamani, Giovanni and Demner-Fushman, Dina and Dligach, Dmitriy and Daneshjou, Roxana and Fernandes, Chrystinne and others},
  title   = {The {TRIPOD-LLM} reporting guideline for studies using large language models},
  journal = {Nature Medicine},
  year    = {2025},
  doi     = {10.1038/s41591-024-03425-5}
}

@article{mehari2022,
  author  = {Mehari, Temesgen and Strodthoff, Nils},
  title   = {Self-supervised representation learning from 12-lead {ECG} data},
  journal = {Computers in Biology and Medicine},
  year    = {2022},
  volume  = {141},
  pages   = {105114}
}

@article{qwen2025,
  author  = {{Qwen Team}},
  title   = {{Qwen3} technical report},
  journal = {arXiv preprint arXiv:2505.09388},
  year    = {2025}
}

@article{llama2024,
  author  = {{Llama Team}},
  title   = {The {Llama} 3 herd of models},
  journal = {arXiv preprint arXiv:2407.21783},
  year    = {2024}
}

@article{deepseek2025,
  author  = {{DeepSeek-AI}},
  title   = {{DeepSeek-R1}: incentivizing reasoning capability in {LLM}s via reinforcement learning},
  journal = {arXiv preprint arXiv:2501.12948},
  year    = {2025}
}

@book{prineas2010,
  author    = {Prineas, Ronald J. and Crow, Richard S. and Zhang, Zhu-Ming},
  title     = {The Minnesota Code Manual of Electrocardiographic Findings},
  edition   = {2},
  publisher = {Springer},
  address   = {London},
  year      = {2010},
  isbn      = {978-1-84882-777-6},
  doi       = {10.1007/978-1-84882-778-3}
}

@article{peguero2017,
  author  = {Peguero, Julio G. and Lo Presti, Saberio and Perez, Jorge and Issa, Omar and Brenes, Juan C. and Tolentino, Alfonso},
  title   = {Electrocardiographic criteria for the diagnosis of left ventricular hypertrophy},
  journal = {Journal of the American College of Cardiology},
  year    = {2017},
  volume  = {69},
  number  = {13},
  pages   = {1694--1703}
}

@article{casale1985,
  author  = {Casale, Paul N. and Devereux, Richard B. and Kligfield, Paul and Eisenberg, Richard R. and Miller, David H. and Chaudhary, Balasubramanian S. and Phillips, Malcolm C.},
  title   = {Electrocardiographic detection of left ventricular hypertrophy: development and prospective validation of improved criteria},
  journal = {Journal of the American College of Cardiology},
  year    = {1985},
  volume  = {6},
  number  = {3},
  pages   = {572--580}
}

@inproceedings{agrawal2022,
  author    = {Agrawal, Monica and Hegselmann, Stefan and Lang, Hunter and Kim, Yoon and Sontag, David},
  title     = {Large language models are few-shot clinical information extractors},
  booktitle = {Proceedings of the 2022 Conference on Empirical Methods in Natural Language Processing},
  year      = {2022},
  pages     = {1998--2022}
}

@inproceedings{irankhah2026archehr,
  title={Yale-DM-Lab at ArchEHR-QA 2026:
         Deterministic Grounding and Multi-Pass Evidence
         Alignment for EHR Question Answering},
  author={Irankhah, Elyas and Fodeh, Samah},
  booktitle={Proceedings of the Third Workshop on
             Patient-Oriented Language Processing
             (CL4Health) at LREC 2026},
  pages={515--523},
  year={2026},
  doi={10.63317/2wd3di8agfag}
}

@article{irankhah2026ragflip,
  title={RAGFlip: Measuring Query-Level Negative Flips
         in Retriever Upgrades},
  author={Irankhah, Elyas and Arif, Muhammad},
  journal={arXiv preprint arXiv:2610.07266},
  year={2026},
  doi={10.48550/arXiv.2610.07266}
}

@article{parklee2024robust,
  title={Toward Robust RALMs: Revealing the Impact of Imperfect Retrieval on Retrieval-Augmented Language Models},
  author={Park, Seong-Il and Lee, Jay-Yoon},
  journal={Transactions of the Association for Computational Linguistics},
  volume={12},
  pages={1686--1702},
  year={2024},
  doi={10.1162/tacl_a_00724}
}

@article{wu2025sourcecheckup,
  title={An Automated Framework for Assessing How Well LLMs Cite Relevant Medical References},
  author={Wu, Kevin and Wu, Eric and Wei, Kevin and Zhang, Angela and Casasola, Allison and Nguyen, Teresa and Riantawan, Sith and Shi, Patricia and Ho, Daniel and Zou, James},
  journal={Nature Communications},
  volume={16},
  pages={3615},
  year={2025},
  doi={10.1038/s41467-025-58551-6}
}

@article{wind2025radiology,
  title={Multi-step Retrieval and Reasoning Improves Radiology Question Answering with Large Language Models},
  author={Wind, Sebastian and Sopa, Jeta and Truhn, Daniel and Lotfinia, Mahshad and Nguyen, Tri-Thien and Bressem, Keno and Adams, Lisa and Rusu, Mirabela and K{\"o}stler, Harald and Wellein, Gerhard and Maier, Andreas and Tayebi Arasteh, Soroosh},
  journal={npj Digital Medicine},
  volume={8},
  pages={790},
  year={2025},
  doi={10.1038/s41746-025-02250-5}
}

\end{document}